%% file: main.tex
\pdfoutput=1

\documentclass[11pt]{article}

\usepackage[final]{acl}

\usepackage{times}
\usepackage{latexsym}

\usepackage[T1]{fontenc}

\usepackage[utf8]{inputenc}

\usepackage{microtype}

\usepackage{inconsolata}

\usepackage{graphicx}

\usepackage{booktabs}

\usepackage{amsmath}

\usepackage{amsfonts}

\title{Assessing the Macro and Micro Effects of Random Seeds on Fine-Tuning Large Language Models}

\author{Nghia T. Bui\textsuperscript{\rm 1}, Guergana Savova\textsuperscript{\rm 2}, Lijing Wang\textsuperscript{\rm 1} \\
        \textsuperscript{\rm 1}New Jersey Institute of Technology \\ \texttt{\{ntb23, lijing.wang\}@njit.edu} \\
        \textsuperscript{\rm 2}Boston Children's Hospital and Harvard Medical School \\ \texttt{guergana.savova@childrens.harvard.edu} \\
        }

\begin{document}
\maketitle

\begin{abstract}
The impact of random seeds in fine-tuning large language models (LLMs) has been largely overlooked despite its potential influence on model performance. 
In this study, we systematically evaluate the effects of random seeds on LLMs using the GLUE and SuperGLUE benchmarks. We analyze the macro impact through traditional metrics like accuracy and F1, calculating their mean and variance to quantify performance fluctuations. To capture the micro effects, we introduce a novel metric, \textit{consistency}, measuring the stability of individual predictions across runs. Our experiments reveal significant variance at both macro and micro levels, underscoring the need for careful consideration of random seeds in fine-tuning and evaluation.
\end{abstract}

\section{Introduction}\label{sec:introduction}
The impact of random seeds in neural network training has long been recognized across various domains, such as general machine learning classification and regression tasks~\cite{ganesh2023impact}~\cite{madhyastha2019model}, computer vision~\cite{picard2021torch}~\cite{aakesson2024random}, natural language processing (NLP)\cite{bethard2022we},\cite{lucic-etal-2022-towards}. 

In the field of NLP, large language models (LLMs) have achieved state-of-the-art results on benchmarks like GLUE and SuperGLUE, which are now standard for evaluating language understanding and reasoning. However, pretrained transformers such as BERT~\cite{devlin-etal-2019-bert} and RoBERTa~\cite{liu2019roberta} are highly sensitive to random seeds~\cite{risch-krestel-2020-bagging, dodge2020fine, mosbach2020stability}, often leading to significant performance variation that complicates experimental interpretation and benchmarking.
While other sources of randomness, such as prompt formatting~\cite{he2024does}, in-context example selection~\cite{gupta2023coverage}, and how learnable weights are initialized~\cite{hayou2024impact}, have also been explored, seed variation remains a fundamental and underaddressed issue.
A recent analysis of 85 papers from the ACL Anthology~\cite{bethard2022we} revealed risky practices in the use of random seeds: over 50\% of the papers exhibited potential misuse, with 24 using a single fixed random seed. \textit{This highlights that random seeds sensitivity in LLM fine-tuning remains insufficiently understood, motivating the need for more systematic investigation.}


\begin{figure}[t]
\centering
\includegraphics[width=1\linewidth]{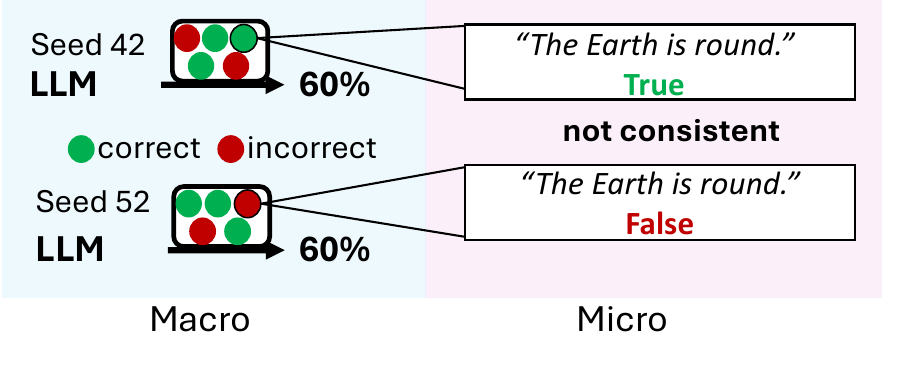}
  \caption{Macro and micro performance. A pretrained LLM is fine-tuned with random seed 42 and 52. The accuracy for both models is 60\%, but the overlapping of individual predictions is 20\%. }
  \label{fig:idea}
\end{figure}


Existing studies examining the impact of random seeds~\cite{ganesh2023impact, madhyastha2019model, picard2021torch} typically evaluate performance variations by measuring the variance of standard metrics, such as accuracy score for classification tasks, or MAE (mean absolute error) for regression tasks, across multiple seeds. These evaluations focus on the \textbf{macro} agreement of model performance across the entire test set, offering insights into overall variability. However, \textit{they overlook the \textbf{micro} impact of how individual test points are influenced by seed-induced variability}. As shown in Figure~\ref{fig:idea}, model performance is robust to random seed 42 and 52 at the macro level (both achieve 60\% accuracy) but lacks consistency at the micro level (only 20\% overlapping predictions). This micro inconsistency can have severe consequences in real-world applications, especially in fields where model predictions are highly sensitive to individual test points, such as medical diagnosis and autonomous driving. Understanding this micro effect is crucial for assessing model robustness at the level of individual predictions, ensuring that specific test samples are not inconsistently predicted due to seed-induced variability. Additionally, it helps pinpoint specific areas where models may exhibit significant instability, such as consistently misclassifying certain types of data points or showing highly variable predictions for similar inputs. Recognizing these areas of instability can guide targeted improvements in both model design and evaluation practices, ensuring that assessments account for seed-induced variability in performance.


\textbf{Major contributions}: 
To address these gaps, in this work, (1) we analyze the impact of random seeds on pretrained LLMs using the GLUE and SuperGLUE benchmarks, covering both macro and micro variability;
(2) We introduce a novel \textbf{consistency} metric to assess prediction stability on individual test points, capturing the micro effects of random seeds;
(3) Our extensive experiments reveal significant variability in both standard and consistency metrics, underscoring the need to consider seed-induced variability in fine-tuning and evaluation, and incorporate random seeds sensitivity into benchmarking and reporting for more reliable and reproducible results.

\section{Macro Metric: Variance}\label{sec:variance}
To measure the macro impact of random seeds on LLM performance, we calculate the variance of a standard evaluation metric across multiple seeds. 
Let \([\zeta_1, \cdots, \zeta_S]\) represent the values of model performance for an LLM fine-tuned with \(S\) random seeds, the \textit{variance} is calculated by:

\begin{equation}
\small
\label{equ:variance}
    \text{VAR}(\zeta) = \sqrt{\frac{1}{S} \sum_{i=1}^S (\zeta_i - \bar{\zeta})^2}
\end{equation}

\noindent where \(\bar{\zeta}=\frac{1}{S} \sum_{i=1}^S \zeta_i\). \(\zeta\) can be any standard metrics, such as accuracy score for classification tasks or MAE for regression tasks. 
A smaller VAR indicates less variation in macro performance.

\section{Micro Metric: Consistency}\label{sec:consistency}

'Consistency' can have varying definitions across domains. Building on prior work, \citet{wang2020wisdom} formally defined the \textit{consistency} of a deep learning model as its ability to produce consistent predictions for the same input when periodically retrained with streaming data in deployment settings. Extending this idea, we define the \textit{consistency} of an LLM as its ability to generate consistent predictions for the same input across models fine-tuned with different hyperparameter settings, with \textit{correct-consistency} further specifying its ability to make consistent correct predictions in this context. 

More specifically, consider two LLMs $A$ and $B$, given a dataset \(\mathcal{D}={d_1,\cdots,d_N}\) of \(N\) data points, $y_i^A$ and $y_i^B$ are the prediction of $A$,$B$ for a data point $d_i$ with ground truth $r_i$. 
The \textit{consistency} of $A$ and $B$ can be calculated as follows:

\begin{equation}
\small
\label{equ:con}
\text{CON:} \frac{1}{N}\sum_{t=1}^{N}{\pi_{A,B}(t)}
\end{equation}

\noindent where \(\pi_{A,B}(\cdot)\) is the scoring function that quantifies the alignment between predictions \(y_t^A\) and \(y_t^B\), with higher values indicating smaller variations in micro-level predictions; it can be either binary (e.g., 1 for match, 0 for mismatch) or probabilistic based on different NLP tasks.
And the \textit{correct-consistency} is calculated by: 

\begin{equation}
\small
\label{equ:ccon}
\text{CCON:} \frac{1}{N}\sum_{t=1}^{N}{\pi_{A,B,r}(t)}
\end{equation}

\noindent where \(\pi_{A,B,r}(\cdot)\) is the scoring function that quantifies the alignment between predictions \(y_t^A\), \(y_t^B\), and ground truth \(r_t\).

In this paper, we focus solely on classification tasks because the GLUE and SuperGLUE benchmarks primarily consist of classification problems. The scoring function \(\pi(\cdot)\) is defined as an indicator function that equals 1 if \(y_t^A = y_t^B (= r_t)\), otherwise 0. 
We summarize the standard metric \(\zeta\) and the possible corresponding scoring functions \(\pi\) for various NLP tasks in Table~\ref{tab:metrics} in the Appendix~\ref{subsec:appex-metric}. We hope this summary offers useful context for interpreting our results and supports future extensions of our evaluation to a broader range of NLP tasks.





 



While consistency metrics can generally be used for quantifying the agreement of individual predictions from any two LLMs with different architectures, hyperparameters, or training settings, in our study, they are specifically used to serve as metrics to evaluate the micro impact of random seeds on the same pretrained LLM.


\input{tables/var-con-all}

\section{Experimental Setup}\label{sec:setup}

\subsection{Benchmarks and pretrained models}\label{subsec:benchmarks}
In this study, we conduct experiments on a range of NLP tasks
including CoLA, SST2, MRPC, QQP, QNLI, and RTE from GLUE~\cite{wang2018glue} benchmark; 
RTE, CB, WiC, BoolQ, MultiRC, and COPA from SuperGLUE~\cite{wang2019superglue} benchmark. 
STSB, WSC, and MNLI tasks are omitted from our experiment. We specify the reason in Section \ref{subsec:appex-sota}. 
All tasks use accuracy as the standard evaluation metric \(\zeta\), except for CoLA, which uses Matthews Correlation Coefficient (MCC).
In our paper, we use RTEG to denote RTE task from GLUE and RTES for SuperGLUE. 

To examine the effects on various scales of LLMs, we experiment with \texttt{\small RoBERTa-large} (\textasciitilde350M trainable parameters), as well as a larger LLM \texttt{\small Llama3.2-3B} (\textasciitilde3.21B trainable parameters) using LoRA~\cite{hu2022lora} fine-tuning, enabling us to assess whether our findings generalize across model scales.


\subsection{Settings}\label{subsec:settings}
Our experiment is implemented using Hugging Face Transformers (v4.30.0) and PyTorch (v2.0), conducted on NVIDIA two A100 GPUs with 80GB of memory each. Based on the empirical findings in ~\cite{wang-etal-2023-two,dodge2020fine,mosbach2020stability}, 5–10 seeds are sufficient to estimate variance of LLMs in NLP tasks. We perform full fine-tuning for each task with ten randomly chosen seeds: 42, 52, 62, 72, 82, 92, 102, 112, 122, 132 (i.e., \(S=10\) ).  We calculate CON and CCON on each unique pair of seeds and report the average of 45 values as the final consistency score. 
Our fine-tuning process is based on the PyTorch script \texttt{run\_glue.py}, and the best previously reported settings were applied unless otherwise specified.

To ensure proper experimental setup and reproducibility, we refer the configurations in~\cite{liu2019roberta} and replicate state-of-the-art (SOTA) scores reported using \texttt{\small RoBERTa-large} with full fine-tuning.
A comparison of our implementation with the reference SOTA scores and detailed data and learning settings are provided in Appendix Table~\ref{tab:benchmarks}, Table~\ref{tab:data}, and Table~\ref{tab:setting}. 
Since we could not find reference configurations or performance reports for \texttt{\small Llama3.2-3B} on the two benchmarks, we use the SOTA scores (Table~\ref{tab:benchmarks}) as a reference and conduct experiments using our own settings, detailed in Table~\ref{tab:llama_setting}.

\section{Results and Discussion}\label{sec:result}

\subsection{Macro impact}\label{subsec:result-macro}
Table~\ref{tab:var-con-all} presents the averaged accuracy (\(\zeta\)) and variance (VAR) for all tasks across ten random seeds using two LLMs. Significant VAR is observed in many tasks using \texttt{\small RoBERTa-large}, such as RTES (16.67), COPA (16.83), and MultiRC (13.30), reflecting sensitivity to random seed selection. 
High variability at the macro level undermines the reliability of single-seed evaluations, emphasizing the need for robust evaluation methods and stability-enhancing techniques. 
In contrast, tasks like QQP (0.06), QNLI (0.38), SST2 (0.55), and MRPC (0.89) show much greater stability, likely due to their inherent properties such as larger datasets or simpler decision boundaries. 
This also helps explain why SuperGLUE tasks generally show higher VAR than GLUE tasks. 

Compared to \texttt{\small RoBERTa-large}, \texttt{\small Llama3.2-3B} with LoRA fine-tuning exhibits significantly lower VAR across most tasks. This is likely because only a small subset of parameters (\textasciitilde2.3 million) is updated during LoRA fine-tuning, which constrains the variance introduced by random seeds and results in greater stability and robustness to seed-induced fluctuations. 
Furthermore, the decoder-only, autoregressive architecture of \texttt{\small Llama3.2-3B} may be inherently less sensitive to minor parameter perturbations during task adaptation compared to the encoder-only structure of \texttt{\small RoBERTa}, contributing to its robust performance.
A deeper analysis of how these factors govern performance variance is a promising avenue for future work.

\subsection{Micro impact}\label{subsec:result-micro}

Table~\ref{tab:var-con-all} reports consistency (CON) and correct-consistency (CCON) for all tasks over ten random seeds. For \texttt{\small RoBERTa-large}, high CON values in tasks like SST2 (96.83), QNLI (95.64), and QQP (95.57) indicate stable predictions, while lower values in tasks like COPA (64.88) and RTES (70.05) highlight their sensitivity to random seeds, potentially due to smaller training sizes or task complexity.
Tasks with large CON–CCON gaps (e.g., WiC with a 21.61 difference and CoLA with 11.19) suggest that \textit{consistent predictions are not always accurate, emphasizing the need to evaluate both stability and correctness}. 
Furthermore, tasks like WiC, which show low VAR alongside low CON and CCON, demonstrate that \textit{similar macro accuracy can mask underlying instability}, reinforcing the importance of micro-level evaluation beyond traditional metrics.
Results from \texttt{\small Llama3.2-3B} show consistent trends, albeit with task-specific variations. 


Unlike the macro impact where \texttt{\small Llama3.2-3B} with LoRA fine-tuning shows significantly lower VAR than \texttt{\small RoBERTa-large}, no such trend is observed for CON and CCON. This implies that 
\textit{using a parameter-efficient method like LoRA with a modern LLM helps mitigate macro variance but has limited effect on micro consistency.}



\subsection{Discussion}\label{subsec:discussion}




\noindent \textbf{\textit{Will increasing training data size improve variance and consistency in general?}}

\begin{figure}[t]
\centering
\includegraphics[width=0.9\linewidth]{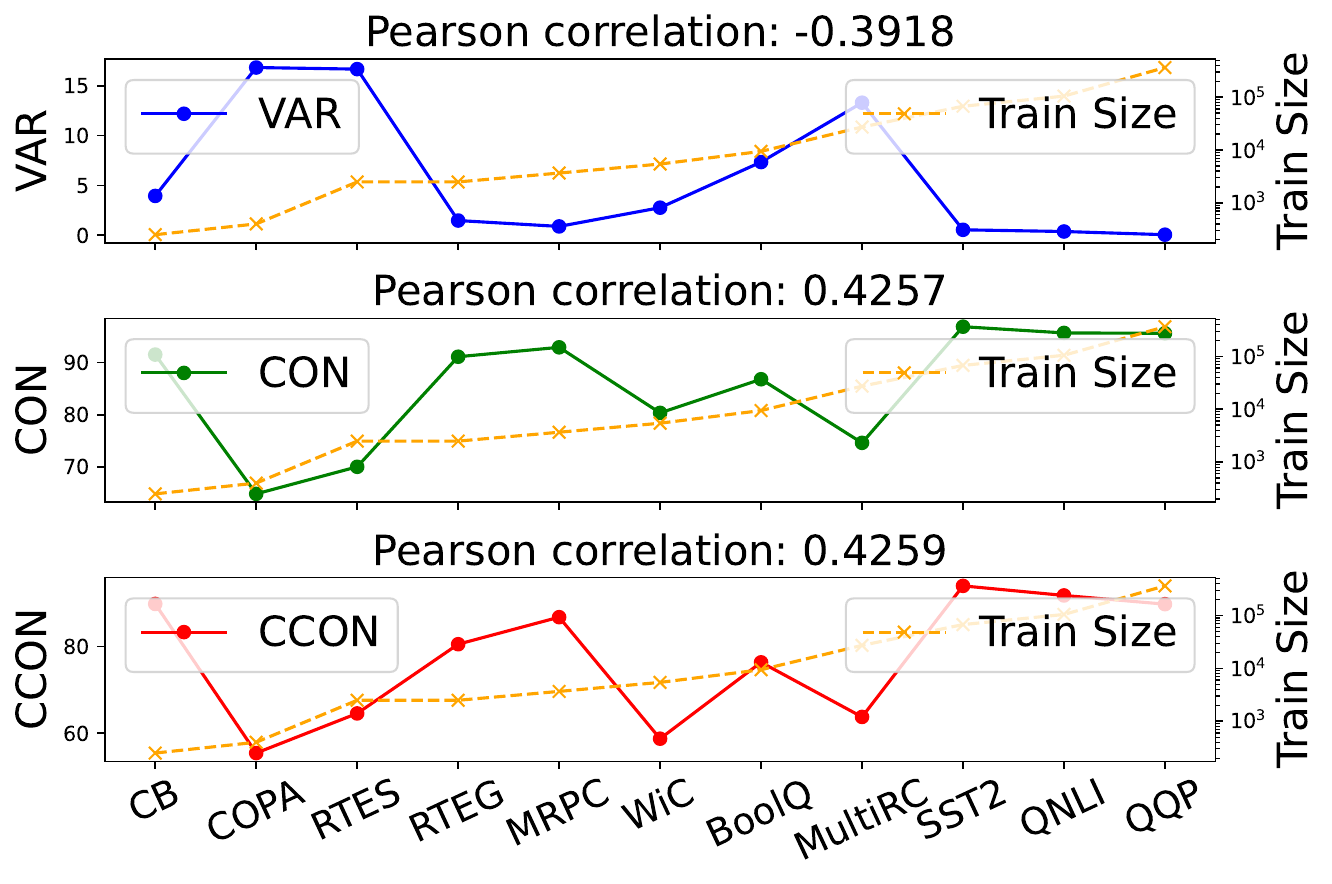}
  \caption{Correlation between training size (log scale), VAR, CON, and CCON while using \texttt{\small RoBERTa-large}. Tasks are arranged in ascending order of training size, with exact sizes detailed in Appendix~\ref{tab:data}. }
  \label{fig:var-size}
\end{figure}

To answer the question, we present Pearson correlation analysis in Figure~\ref{fig:var-size}, showing the relationship between training size, variance, and consistency, with tasks sorted by increasing dataset size. It reveals a weak negative correlation (-0.3918) between training size and VAR, indicating that smaller datasets tend to increase macro variance. 
However, the effect is not pronounced or consistent across all tasks, as MultiRC exhibits high VAR despite a relatively large dataset.
A weak or moderate positive correlation is observed between training size and both CON (0.4257) and CCON (0.4259), suggesting that larger datasets generally improve consistency and prediction stability across random seeds, but with no guarantee. 
\textit{Increasing training size can reduce macro and micro variability in random seeds, but its effectiveness depends on factors like data quality, task complexity, and label noise~\cite{shahinfar2020many,althnian2021impact,bailly2022effects}.} 


\noindent \textit{\textbf{Outlook: mitigating seed-induced variability.}}

Our analysis reveals substantial seed-induced variability in model performance at both the macro and micro levels. The findings indicate that increasing the training size can reduce both macro and micro variability in random seeds, but it’s not a guaranteed solution, especially for complex tasks.

Prior studies have proposed several strategies to mitigate macro variability, including model ensembling~\cite{risch-krestel-2020-bagging, wang-etal-2023-two, wang2020wisdom}, stability-aware training~\cite{dodge2020fine}, and more robust evaluation protocols~\cite{mosbach2020stability}.
Among these methods, only \citet{wang2020wisdom} explicitly addresses consistency through snapshot ensembles. While this method guarantees improvement, it is computationally prohibitive for LLMs, as it requires training numerous models to form a sufficient ensemble.
 

Inspired by our observation that individual predictions are highly sensitive to randomness, a promising future direction is the development of novel optimization algorithms. Specifically, one could dynamically weight the loss of individual examples during gradient propagation to stabilize training and reduce both macro and micro variance, without incurring significant computational overhead. We leave the exploration of such methods for future work.

\section{Conclusion}\label{sec:conclusion}
In conclusion, this work highlights the significant impact of random seeds on pretrained LLMs, revealing variability at both macro and micro levels. By introducing a novel consistency metric, we emphasize the importance of considering seed-induced variability in individual predictions in model evaluation. Our findings stress the need for incorporating random seed sensitivity into benchmarking for more reliable and reproducible results.

\section{Limitations}\label{sec:limitation}
Our work focuses on classification tasks which are the main consists of GLUE and SuperGLUE benchmarks. While this provides a solid foundation, it may limit the generalizability of our findings to other NLP task types. Incorporating a broader set of benchmark datasets would allow for a more comprehensive evaluation using our proposed macro and micro metrics across diverse task categories (as summarized in Table~\ref{tab:metrics}). Experimenting with greater task diversity would better capture variability in model behavior, ultimately enhancing the robustness and applicability of our analysis. We leave this broader exploration as future work.

\section*{Acknowledgement}
The authors would like to thank the anonymous reviewers for feedback that improved the paper, the New Jersey Institute of Technology (NJIT) and the US National Institutes of Health (NIH)  for providing funding. This research is supported by NJIT New Faculty Grant and NIH Grant R01LM013486. Any opinions, findings, and conclusions or recommendations expressed in this material are those of the authors and do not necessarily reflect the views of NJIT or NIH.

\bibliography{main}

\appendix

\section{Appendix}
\label{sec:appendix}

\subsection{Macro and Micro Metrics for Various NLP Tasks}\label{subsec:appex-metric}

Given two LLMs $A$ and $B$, a dataset \(\mathcal{D}={d_1,\cdots,d_N}\) of \(N\) data points, $y_i^A$ and $y_i^B$ are the prediction of $A$,$B$ for a data point $d_i$ with ground truth $r_i$. 
Table~\ref{tab:metrics} summarizes the performance metric \(\zeta\) and the corresponding scoring functions 
\(\pi(\cdot)\) used in Equations~\ref{equ:con} and~\ref{equ:ccon} for various NLP tasks.

\subsection{Data Description}\label{subsec:appex-data}
Table~\ref{tab:data} presents the statistics of the dataset used in our experiments. Each dataset consists of predefined train, dev and test data in CSV format. We use the train and dev sets for training and evaluation. Since the test set does not include gold-standard labels, the dev set also serves as the test set. For datasets where each instance may have multiple correct answers, such as MultiRC, we split the data at the question-answer pair level rather than the passage level. This ensures a more balanced distribution of instances across the train and dev sets.
In the COPA dataset, each instance is originally described by six fields—premise, choice1, choice2, question, idx, and label. To adapt these instances into a multiple-choice format, we construct two candidate sequences for every sample. Specifically, for each candidate, we concatenate the premise with the question and the corresponding choice using a dedicated separation token (e.g., “[SEP]”) to clearly delineate the different textual components. We then maintain the original label field, converting it from 1/2 to 0/1 to match the 0-based index convention in multiple-choice classification.  This preprocessing ensures consistency with other classification tasks and allows the model to effectively learn the relationships between the premise and possible choices.

\subsection{Hyperparameter Settings}\label{subsec:appex-hyper}
Table~\ref{tab:setting} and Table~\ref{tab:llama_setting} provides the detailed hyperparameter configurations. Unless stated otherwise, we adopt the default hyperparameter values from the Hugging Face framework.

\subsection{Replicated SOTA Scores}\label{subsec:appex-sota}
To ensure the reproducibility of our experiments in SuperGLUE and GLUE tasks, we adhered to the specified settings and reproduced the state-of-the-art (SOTA) accuracy scores reported in:
\url{https://github.com/facebookresearch/fairseq/tree/main/examples/roberta}.
Although differences in the fine-tuning script and missing settings from the original authors prevented us from reproducing the exact SOTA scores, our replicated accuracy scores for the GLUE and SuperGLUE tasks, presented in Table~\ref{tab:benchmarks}, are directly comparable and align with those shown in Table~\ref{tab:var-con-all} of the main paper and Table~\ref{tab:roberta-all} in Section~\ref{subsec:appex-result}. 

We were unable to replicate the representation (special token extractions) and model settings (unpublished pretrained model) for the WSC and MNLI tasks, so they are omitted from the experiment.
STSB is a regression task, thus is omitted. 

\subsection{Additional Results}\label{subsec:appex-result}
Table~\ref{tab:roberta-all} and Table~\ref{tab:llama-all} present model performance across various metrics, including precision (P), recall (R), F1 score, accuracy, CON, and CCON, with average values and standard deviations (VAR). In Section~\ref{sec:result} of the main paper, significant variance in macro-level performance across many tasks highlights sensitivity to random seed selection. Similar patterns in the VAR values for P, R, and F1 further confirm the robustness of our findings across various standard metrics.

\subsection{Additional Discussion}
\label{subsec:appex-discussion}

\noindent \textit{\textbf{Is there a universally superior random seed?} }

\begin{figure}[ht]
\centering
\includegraphics[width=0.9\linewidth]{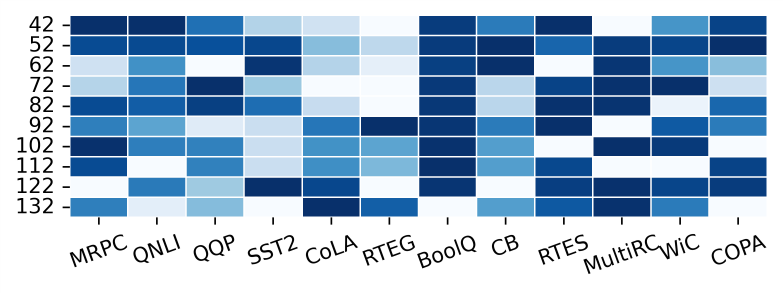}
  \caption{A heatmap of normalized \(\zeta\) values across tasks and ten random seeds, with a darker color representing a better accuracy. 
  }
  \label{fig:heatmap}
\end{figure}

To investigate whether a specific random seed consistently leads to better results across different models or tasks, in Figure~\ref{fig:heatmap} we present a heatmap of normalized accuracy for each task across ten random seeds using \texttt{\small RoBERTa-large}. There is no significant difference in color distribution between each row, indicating that \textit{no discernible pattern or evidence supporting the existence of a universally superior random seed}.

\noindent \textbf{\textit{Will increasing training data size improve variance and consistency in general?}}

\begin{figure}[!t]
\centering
\includegraphics[width=0.9\linewidth]{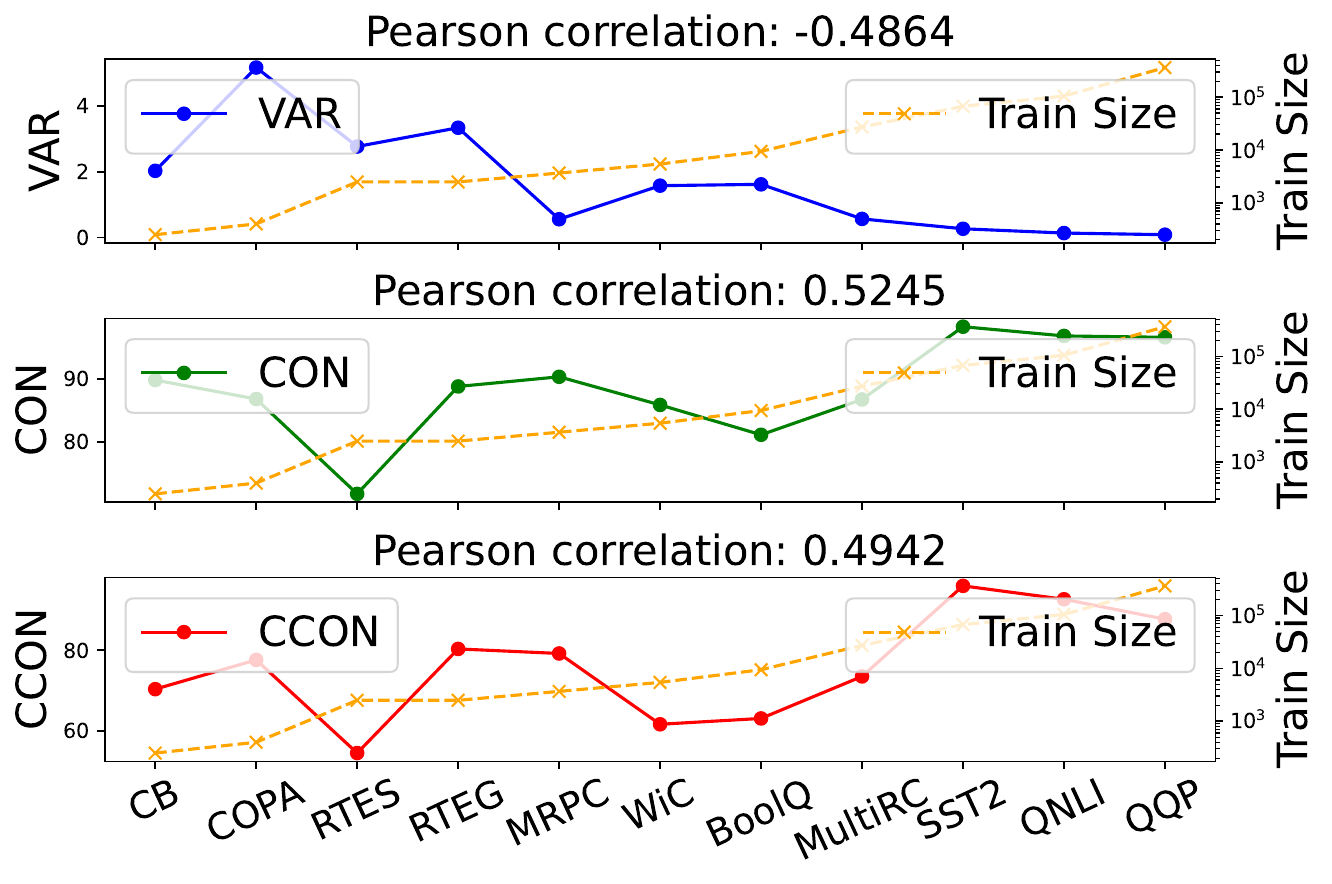}
  \caption{Correlation between training size (log scale), VAR, CON, and CCON while using \texttt{\small Llama3.2-3B}. Tasks are arranged in ascending order of training size, with exact sizes detailed in Appendix~\ref{tab:data}.}
  \label{fig:var-size_llama}
\end{figure}

Figure~\ref{fig:var-size_llama} presents the correlation analysis of VAR, CON, and CCON using \texttt{\small Llama3.2-3B}, revealing patterns consistent with the findings discussed in Section~\ref{subsec:discussion}.

\input{tables/score_functions}
\input{tables/benchmarks}

\input{tables/data}

\input{tables/setting}

\input{tables/llama_setting}

\input{tables/results_roberta_10_seeds}
\input{tables/llama_results}
\end{document}

%% file: tables/var-con-all.tex
\begin{table*}
\small
\centering
\setlength{\tabcolsep}{0.26em}
\begin{tabular}{lcccccclcccccc}
\toprule
\multicolumn{7}{c}{\texttt{\small{RoBERTa-large}}} & \multicolumn{7}{c}{\texttt{\small{Llama3.2-3B}}} \\
\cmidrule(r){1-7}\cmidrule(r){8-14}
\texttt{\small{GLUE}} & \textbf{MRPC} & \textbf{QNLI} & \textbf{QQP} & \textbf{SST2} & \textbf{RTEG} & \textbf{CoLA} & \texttt{\small{GLUE}} & \textbf{MRPC} & \textbf{QNLI} & \textbf{QQP} & \textbf{SST2} & \textbf{RTEG} & \textbf{CoLA} \\
\cmidrule(r){1-7}\cmidrule(r){8-14}

\(\zeta\)	&	90.34&	94.00&	92.00&	95.59&  85.02&	65.61& \(\zeta\) &	84.02&	94.20&	89.37&	96.78&	85.92&	61.33\\
VAR	&	0.89&	0.38&	0.06&	0.55&  1.48&	1.32& VAR	&	0.56&	0.14&	0.09&	0.27&	3.34&	0.88\\
CON	&	92.89&	95.64&	95.57&	96.83&  91.09&	93.95& CON	&	90.34&	96.85&	96.63&	98.32&	88.80&	94.86\\
CCON	&	86.79&	91.80&	89.79&	94.01&  80.55&	82.76& CCON	&	79.19&	92.63&	87.69&	95.95&	80.32&	81.36\\

\cmidrule(r){1-7}\cmidrule(r){8-14}

\texttt{\small{SuperGLUE}} & \textbf{BoolQ} & \textbf{CB} & \textbf{RTES} & \textbf{MultiRC} & \textbf{WiC} & \textbf{COPA} & \texttt{\small{SuperGLUE}} & \textbf{BoolQ} & \textbf{CB} & \textbf{RTES} & \textbf{MultiRC} & \textbf{WiC} & \textbf{COPA} \\
\cmidrule(r){1-7}\cmidrule(r){8-14}

\(\zeta\)	&	83.05&	94.64&	72.89&	76.46&  68.57&	73.00& \(\zeta\)	&	72.49&	73.92&	68.66&	80.14&	68.71&	84.20\\
VAR	&	7.35&	3.95&	16.67&	13.30&  2.77&	16.83& VAR	&	1.62&	2.03&	2.77&	0.57&	1.58&	5.17\\
CON	&	86.79&	91.50&	70.05&	74.63&  80.36&	64.88& CON	&	81.10&	89.82&	71.69&	86.73&	85.86&	86.8\\
CCON	&	76.37&	89.84&	64.58&	63.77&  58.75&	55.44& CCON	&	63.09&	70.35&	54.51&	73.50&	61.64&	77.6\\
\bottomrule
\end{tabular}
\caption{Macro and micro impact of ten random seeds. \(\zeta\) is the average of 10 values, which is MCC for CoLA and accuracy for the other tasks. VAR is the variance of \(\zeta\) calculated using Equation~\ref{equ:variance}. CON and CCON are the average of 45 consistency values calculated using Equation~\ref{equ:con} and \ref{equ:ccon}.
\(\zeta\), CON, and CCON are expressed as percentages. 
}
\label{tab:var-con-all}
\end{table*}

%% file: tables/score_functions.tex
\begin{table*}
\small
\centering
\setlength{\tabcolsep}{0.20em}
\renewcommand{\arraystretch}{1.4}
\begin{tabular}{l|l|l|l}
\toprule
\textbf{NLP Task Type} & \textbf{Standard Metric \(\zeta\)} & \textbf{Scoring Function \(\pi(y_i^A, y_i^B)\)} & \textbf{Scoring Function \(\pi(y_i^A, y_i^B, r_i)\)} \\ \hline

Classification &
Accuracy, F1 &
\(\mathbf{1}(y_i^A = y_i^B)\) &
\(\mathbf{1}(y_i^A = y_i^B = r_i)\) \\ \hline

Regression &
Pearson/Spearman, MSE, MAE &
\(\zeta(y^A, y^B)\) &
\(\frac{1}{2}(\zeta(y_i^A, r_i) + \zeta(y_i^B, r_i))\) \\ \hline

Text Generation &
BLEU, ROUGE, BERTScore &
\(\zeta(y_i^A, y_i^B)\) &
\(\frac{1}{2}(\zeta(y_i^A, r_i) + \zeta(y_i^B, r_i))\) \\ \hline

Sequence Labeling &
Token-/Span-level F1 &
\(\frac{1}{T} \sum_t \mathbf{1}(y_{i,t}^A = y_{i,t}^B)\) &
\(\frac{1}{T} \sum_t \mathbf{1}(y_{i,t}^A = y_{i,t}^B = r_{i,t})\) \\ \hline

Structured Prediction &
UAS, LAS, Exact Match &
\(\zeta(y_i^A, y_i^B)\) &
\(\frac{1}{2}(\zeta(y_i^A, r_i) + \zeta(y_i^B, r_i))\) \\ \hline

Question Answering &
Exact Match, F1 (token overlap) &
\(\zeta(y_i^A, y_i^B)\) &
\(\frac{1}{2}(\zeta(y_i^A, r_i) + \zeta(y_i^B, r_i))\) \\ \hline

Retrieval / Ranking &
NDCG, MRR, MAP &
\(\zeta(y_i^A, y_i^B)\) &
\(\frac{1}{2}(\zeta(y_i^A, r_i) + \zeta(y_i^B, r_i))\) \\

\bottomrule
\end{tabular}
\caption{Common evaluation metrics and scoring functions for various NLP task types. \(\mathbf{1}(\cdot)\) denotes the indicator function that equals 1 or 0. 
For generation and ranking tasks, similarity and ranking correlation metrics (like BERTScore or Kendall Tau) are more appropriate.
Note that this table includes only a subset of commonly used \(\zeta\) metrics; many other relevant metrics are omitted due to space limitations.}
\label{tab:metrics}
\end{table*}

%% file: tables/benchmarks.tex


\begin{table*}
\small
\centering
\setlength{\tabcolsep}{0.18em}
\begin{tabular}{lccccccc}
\toprule
\texttt{\small{GLUE}} & \textbf{MRPC} & \textbf{QNLI} & \textbf{QQP} & \textbf{SST2} & \textbf{RTEG} & \textbf{CoLA} \\
\midrule
Reference & 90.9 & 94.7 & 92.2 & 96.4 & 86.6 & 68.0 \\
Replicated & 91.2 & 94.7 & 92.1 & 96.9 & 84.8 & 65.3 \\
\midrule
\texttt{\small{SuperGLUE}} & \textbf{BoolQ} & \textbf{CB} & \textbf{RTES} & \textbf{MultiRC} & \textbf{WiC} & \textbf{COPA} \\
\midrule
Reference & 86.9 & 98.2 & 89.5 & 85.7 & 75.6 & 94.0  \\
Replicated & 85.4 & 100 & 86.3 & 84.9 & 71.2 & 90.0  \\
\bottomrule
\end{tabular}
\caption{Reference and replicated scores on the GLUE and SuperGLUE tasks. These scores are obtained by training on the train set, validating and testing on the dev set.}
\label{tab:benchmarks}
\end{table*}

%% file: tables/data.tex
\begin{table*}[]
\small
\centering
\setlength{\tabcolsep}{0.24em}
\begin{tabular}{lcccccc}
\toprule
\texttt{\small GLUE}      & \textbf{MRPC}  & \textbf{QNLI} & \textbf{QQP}  & \textbf{SST2}    & \textbf{RTEG} & \textbf{CoLA} \\ \midrule
Classes                                              & 2              & 2             & 2             & 2                & 2             & 2             \\
Train samples                                        & 3668           & 104743        & 363846        & 67349            & 2490          & 8551          \\
Dev samples                                          & 408            & 5463          & 40430         & 872              & 277           & 1043          \\
Test samples                                         & 1725           & 5463          & 390965        & 1821             & 3000          & 1063          \\ \midrule
\texttt{\small SuperGLUE} & \textbf{BoolQ} & \textbf{CB}   & \textbf{RTES} & \textbf{MultiRC} & \textbf{WiC}  & \textbf{COPA} \\ \midrule
Classes                                              & 2              & 3             & 2             & 2                & 2             & 2             \\
Train samples                                        & 9427           & 250           & 2490          & 27243            & 5428          & 400           \\
Dev samples                                          & 3270           & 56            & 277           & 4848             & 638           & 100           \\
Test samples                                         & 3245           & 250           & 3000          & 9693             & 1400          & 500           \\ \bottomrule
\end{tabular}
\caption{Data statistics for GLUE and SuperGLUE.}
\label{tab:data}
\end{table*}

%% file: tables/setting.tex
\begin{table*}
\small
\centering
\setlength{\tabcolsep}{0.24em}
\begin{tabular}{lccccccc}
\toprule
\texttt{\small{GLUE}} & \textbf{MRPC} & \textbf{QNLI} & \textbf{QQP} & \textbf{SST2} & \textbf{RTEG} & \textbf{CoLA} \\ 
\midrule
Random seed & 42 & 72 & 42 & 52 & 52 & 72\\
Batch size & 10 & 10 & 10 & 10 & 10 & 10\\
Epoch & 8 & 6 & 8 & 7 & 10 & 8\\
Learning rate & 2e-5 & 2e-5 & 1e-5 & 2e-5 & 1e-5 & 1e-5\\
Learning rate schedule type & \textit{linear} & \textit{linear} & \textit{linear} & \textit{linear} & \textit{linear} & \textit{linear}\\
Max sequence length & 512 & 512 & 512 & 512 & 512 & 512\\
Gradient accumulation steps & 2 & 2 & 2 & 2 & 2 & 2\\
\midrule
\texttt{\small{SuperGLUE}} & \textbf{BoolQ} & \textbf{CB} & \textbf{RTES} & \textbf{MultiRC} & \textbf{WiC} & \textbf{COPA} \\
\midrule
Random seed & 62 & 52 & 72 & 72 & 42 & 52 \\
Batch size & 10 & 10 & 10 & 10 & 10 & 10 \\
Epoch & 8 & 7 & 10 & 6 & 8 & 9 \\
Learning rate & 1e-5 & 2e-5 & 2e-5 & 2e-5 & 1e-5 & 3e-5 \\
Learning rate schedule type & \textit{linear} & \textit{linear} & \textit{linear} & \textit{linear} & \textit{linear} & \textit{linear} \\
Max sequence length & 512 & 512 & 512 & 512 & 512  & 256\\
Gradient accumulation steps & 2 & 2 & 2 & 2 & 2 & 2 \\
\bottomrule
\end{tabular}
\caption{The hyperparameter settings for GLUE and SuperGLUE tasks to replicate the reference performance in Table~\ref{tab:benchmarks} using \texttt{\small RoBERTa-large}.}
\label{tab:setting}
\end{table*}

%% file: tables/llama_setting.tex
\begin{table*}[]
\centering
\small
\setlength{\tabcolsep}{0.24em}
\begin{tabular}{lccccccc}
\toprule
\texttt{\small{GLUE}}       & \textbf{MRPC}   & \textbf{QNLI}   & \textbf{QQP}    & \textbf{SST2}    & \textbf{RTEG}   & \textbf{CoLA}  \\
\midrule
Batch size                  & 10     & 10     & 32     & 10      & 10     & 10      \\
Epoch                       & 20     & 6      & 2      & 6       & 20     & 6       \\
Learning rate               & 3e-5   & 3e-5   & 2e-5   & 3e-5    & 3e-5   & 1e-5    \\
Learning rate schedule type & linear & linear & linear & linear  & linear & linear  \\
Max sequence length         & 512    & 512    & 512    & 512     & 512    & 512     \\
Gradient accumulation steps & 2      & 2      & 2      & 2       & 2      & 2       \\
LoRA $r_q, r_v$             & 8      & 8      & 8      & 8       & 8      & 8       \\
LoRA $\alpha$               & 16     & 16     & 16     & 16      & 16     & 16      \\
LoRA Dropout                & 0.1    & 0.1    & 0.1    & 0.1     & 0.1    & 0.1      \\
\midrule
\texttt{\small{SuperGLUE}}  & \textbf{BoolQ}  & \textbf{CB}     & \textbf{RTES}   & \textbf{MultiRC} & \textbf{WiC}    & \textbf{COPA}           \\
\midrule
Batch size                  & 10     & 10     & 10     & 10      & 10     & 10      \\
Epoch                       & 10     & 20     & 10     & 10      & 10     & 10      \\
Learning rate               & 1e-5   & 5e-5   & 1e-5   & 2e-5    & 1e-5   & 3e-5    \\
Learning rate schedule type & linear & linear & linear & linear  & linear & linear  \\
Max sequence length         & 512    & 512    & 512    & 512     & 512    & 256     \\
Gradient accumulation steps & 2      & 2      & 2      & 2       & 2      & 2       \\
LoRA $r_q, r_v$             & 8      & 8      & 8      & 8       & 8      & 8       \\
LoRA $\alpha$               & 16     & 16     & 16     & 16      & 16     & 16      \\
LoRA Dropout                & 0.1    & 0.1    & 0.1    & 0.1     & 0.1    & 0.1      \\
\bottomrule
\end{tabular}
\caption{The hyperparameters settings for GLUE and SuperGLUE tasks for \texttt{\small Llama3.2-3B}. For the COPA dataset we implement \texttt{\small LlamaForMultipleChoice} by ourselves.}
\label{tab:llama_setting}
\end{table*}

%% file: tables/results_roberta_10_seeds.tex
\begin{table*}
\small
\centering
\setlength{\tabcolsep}{0.20em}
\begin{tabular}{lcccccclcccccc}
\toprule
 & \multicolumn{6}{c}{\textbf{GLUE}} & & \multicolumn{6}{c}{\textbf{SuperGLUE}} \\
\cmidrule(r){1-7}\cmidrule(r){8-14}
\textbf{Tasks} & P & R & F1  & Accuracy & CON & CCON & \textbf{Tasks} & P & R & F1  & Accuracy & CON & CCON \\
\midrule
MRPC  & 89.62& 87.83& 88.59& 90.34        & 92.89        & 86.79        & BoolQ   & 79.09& 80.78& 79.71& 83.05         & 86.79         & 76.37         \\
      & ($\pm$1.39)& ($\pm$1.15)& ($\pm$1.02)& ($\pm$ 0.89) & ($\pm$ 1.18) & ($\pm$ 0.91) &         & ($\pm$16.87)& ($\pm$10.82)& ($\pm$14.54)& ($\pm$ 7.35)  & ($\pm$ 12.21) & ($\pm$ 10.76) \\
QNLI  & 93.98& 93.98& 93.97& 94.00        & 95.64        & 91.80        & CB      & 89.43& 90.81& 89.61& 94.64& 91.50& 89.84\\
      & ($\pm$0.36)& ($\pm$0.36)& ($\pm$0.36)& ($\pm$ 0.38) & ($\pm$ 0.52) & ($\pm$ 0.46) &         & ($\pm$ 6.15)& ($\pm$ 6.69)& ($\pm$ 6.11)& ($\pm$ 3.95)& ($\pm$ 3.21)& ($\pm$ 3.40)\\
QQP   & 91.22& 91.73& 91.46& 92.00        & 95.57        & 89.79        & RTES    & 71.71& 76.49& 72.99& 72.89         & 70.05& 64.58\\
      & ($\pm$0.07)& ($\pm$0.17)& ($\pm$0.07)& ($\pm$ 0.06) & ($\pm$ 0.30) & ($\pm$ 0.16) &         & ($\pm$25.46)& ($\pm$14.21)& ($\pm$21.72)& ($\pm$ 16.67) & ($\pm$ 18.36)& ($\pm$17.43)\\
SST2  &              95.48&              95.45&              95.46& 95.59        & 96.83        & 94.01        & MultiRC &               67.62&               74.32&               70.05&               76.46&               74.63&               63.77\\
      & ($\pm$0.42)& ($\pm$0.43)& ($\pm$0.43)& ($\pm$ 0.55) & ($\pm$ 0.57) & ($\pm$ 0.51) &         & ($\pm$26.94)& ($\pm$16.79)& ($\pm$23.24)& ($\pm$13.30)& ($\pm$18.94)& ($\pm$15.65)\\
RTEG  & 85.41& 84.74& 84.86& 85.02        & 91.09        & 80.55        & WiC     & 0.7053& 0.6857& 0.6778& 68.57         & 80.36         & 58.75         \\
      & ($\pm$1.45)& ($\pm$1.54)& ($\pm$1.52)& ($\pm$ 1.48) & ($\pm$ 1.74) & ($\pm$ 1.34) &         & ($\pm$2.68)& ($\pm$2.77)& ($\pm$3.28)& ($\pm$ 2.77)  & ($\pm$ 6.07)  & ($\pm$ 4.60)  \\
CoLA  & -& -& -& 65.61        & 93.95        & 82.76        & COPA    & 73.03& 72.85& 72.78& 73.00& 64.88         & 55.44         \\
      & -& -& -& ($\pm$ 1.32) & ($\pm$ 0.71) & ($\pm$ 0.51) &         & ($\pm$ 16.69)& ($\pm$ 16.59)& ($\pm$ 16.81)& ($\pm$ 16.83)& ($\pm$ 17.35) & ($\pm$ 11.98) \\
\bottomrule
\end{tabular}
\caption{Additional evaluation performance using \texttt{\small RoBERTa-large}. P - Precision, R - Recall, F1, Accuracy, CON - consistency, CCON - correct consistency. CoLA employs Matthew’s correlation coefficient as the accuracy.}
\label{tab:roberta-all}
\end{table*}


%% file: tables/llama_results.tex
\begin{table*}
\small
\centering
\setlength{\tabcolsep}{0.20em}
\begin{tabular}{lcccccclcccccc}
\toprule
 & \multicolumn{6}{c}{\textbf{GLUE}} & & \multicolumn{6}{c}{\textbf{SuperGLUE}} \\
\cmidrule(r){1-7}\cmidrule(r){8-14}
\textbf{Tasks} & P & R & F1  & Accuracy & CON & CCON & \textbf{Tasks} & P & R & F1  & Accuracy & CON & CCON \\
\midrule
MRPC  &   81.98&   80.31&    81.03&           84.02&     90.34&         79.19& BoolQ   &   70.78&   69.29&    69.74& 72.49&     81.10&         63.09\\
      &   ($\pm$ 0.57)&   ($\pm$1.01)&    ($\pm$0.79)&           ($\pm$ 0.56)&     ($\pm$1.66)&         ($\pm$0.70)&         &   ($\pm$1.77)&   ($\pm$2.07)&    ($\pm$2.07)& ($\pm$ 1.62) &     ($\pm$1.67)&         ($\pm$1.58)\\
QNLI  &   94.22&   94.22&    94.21& 94.20     &     96.85&         92.63& CB      &   56.33&   56.23&    55.32& 73.92&     89.82&         70.35\\
      &   ($\pm$0.15)&   ($\pm$0.15)&    ($\pm$0.14)& ($\pm$ 0.14) &     ($\pm$0.24)&         ($\pm$0.18)&         &   ($\pm$10.07)&   ($\pm$3.77)&    ($\pm$5.78)& ($\pm$ 2.03)&     ($\pm$2.92)&         ($\pm$2.55)\\
QQP   &   88.36&   89.08&    88.69& 89.37&     96.63&         87.69& RTES    &   68.83&   68.82&    68.64& 68.66     &     71.69&         54.51\\
      &   ($\pm$0.07)&   ($\pm$0.21)&    ($\pm$0.12)& ($\pm$ 0.09) &     ($\pm$0.20)&         ($\pm$0.09)&         &   ($\pm$2.84)&   ($\pm$2.83)&    ($\pm$2.78)& ($\pm$ 2.77) &     ($\pm$3.82)&         ($\pm$3.38)\\
SST2  &   96.78&   96.79&    96.78& 96.78&     98.32&         95.95& MultiRC &   79.72&   80.02&    79.83&           80.14&     86.73&         73.50\\
      &   ($\pm$0.27)&   ($\pm$0.27)&    ($\pm$0.27)& ($\pm$0.27)&     ($\pm$0.63)&         ($\pm$0.44)&         &   ($\pm$0.57)&   ($\pm$0.55)&    ($\pm$0.57)&           ($\pm$0.57)&     ($\pm$1.06)&         ($\pm$0.84)\\
RTEG  &   85.92&   85.91&    85.88& 85.92     &     88.80&         80.32& WiC     &   68.99&   68.71&    68.60& 68.71     &     85.86&         61.64\\
      &   ($\pm$3.36)&   ($\pm$3.34)&    ($\pm$3.35)& ($\pm$ 3.34) &     ($\pm$3.05)&         ($\pm$3.40)&         &   ($\pm$1.64)&   ($\pm$1.58)&    ($\pm$1.58)& ($\pm$ 1.58) &     ($\pm$1.79)&         ($\pm$1.08)\\
CoLA  &   -&   -&    -& 61.33     &     94.86&         81.36& COPA    &   84.08&   84.05&    84.03&           84.20&     86.8&         77.6\\
      &   -&   -&    -& ($\pm$ 0.88) &     ($\pm$0.56)&         ($\pm$ 0.33)&         &   ($\pm$5.25)&   ($\pm$5.28)&    ($\pm$5.24)&           ($\pm$5.17)&     ($\pm$ 3.99)&         ($\pm$ 4.83)\\
\bottomrule
\end{tabular}
\caption{Additional evaluation performance using \texttt{\small Llama3.2-3B}. P - Precision, R - Recall, F1, Accuracy, CON - consistency, CCON - correct consistency. CoLA employs Matthew’s correlation coefficient as the accuracy.}
\label{tab:llama-all}
\end{table*}